\documentclass{article}
\usepackage{microtype}
\usepackage{graphicx}
\usepackage{subfigure}
\usepackage{booktabs} 

\usepackage{arydshln}
\usepackage[dvipsnames,table,xcdraw]{xcolor}
\usepackage{soul}
\usepackage{url}
\usepackage{multirow}
\usepackage{float}
\usepackage{dblfloatfix}
\usepackage{wrapfig}
\usepackage{capt-of}
\usepackage{enumitem}
\usepackage{pdfpages}

\newcommand{\ignore}[1]{}
\newcommand\B{\rule[-1.2ex]{0pt}{0pt}} 

\newcommand\Te{\rule{0pt}{2.5ex}}       
\newcommand\Be{\rule[-2.0ex]{0pt}{0pt}} 

\newcommand{\parahead}[1]{\textbf{#1}:\ }

\newenvironment{packed_itemize}
{\begin{itemize}[leftmargin=0.3cm]
    \setlength{\itemsep}{1pt}
    \setlength{\parskip}{0pt}
    \setlength{\parsep}{0pt}
}{\end{itemize}}

\usepackage{hyperref}

\usepackage[accepted]{icml2019}

\icmltitlerunning{Implicit Filter Sparsification in Convolutional Neural Networks}

\begin{document}

\twocolumn[
\icmltitle{Implicit Filter Sparsification In Convolutional Neural Networks}



\icmlsetsymbol{equal}{*}

\begin{icmlauthorlist}
\icmlauthor{Dushyant Mehta}{mpi,saar}
\icmlauthor{Kwang In Kim}{unist}
\icmlauthor{Christian Theobalt}{mpi}
\end{icmlauthorlist}

\icmlaffiliation{mpi}{Max Planck Institute For Informatics, Saarbr\"ucken, Germany}
\icmlaffiliation{unist}{Ulsan National Institute of Science and Technology, South Korea}
\icmlaffiliation{saar}{Saarland Informatics Campus, Germany}

\icmlcorrespondingauthor{Dushyant Mehta}{dmehta@mpi-inf.mpg.de}

\icmlkeywords{Machine Learning, ICML}

\vskip 0.3in
]



\printAffiliationsAndNotice{}  

\begin{abstract}
We show implicit \emph{filter level} sparsity manifests in convolutional neural networks (CNNs) which employ Batch Normalization and ReLU activation, and are trained with adaptive gradient descent techniques and L2 regularization or weight decay. Through an extensive empirical study \cite{anonymous} we hypothesize the mechanism behind the sparsification process, and find surprising links to certain filter sparsification heuristics proposed in literature. Emergence of, and the subsequent pruning of selective features is observed to be one of the contributing mechanisms, leading to feature sparsity at par or better than certain explicit sparsification / pruning approaches. In this workshop article we summarize our findings, and point out corollaries of selective-feature-penalization which could also be employed as heuristics for filter pruning.
\end{abstract}

\section{Introduction}
In this article we discuss the findings from \cite{anonymous} regarding filter level sparsity which emerges in certain types of feedforward convolutional neural networks. Filter refers to the weights and the nonlinearity associated with a particular feature, acting together as a unit. We use filter and feature interchangeably throughout the document. We particularly focus on the implications of the findings on feature pruning for neural network speed up. 

In networks which employ Batch Normalization and ReLU activation, after training, certain filters are observed to not activate for any input. Importantly, the sparsity emerges in the presence of regularizers such as L2 and weight decay (WD) which are in general understood to be non sparsity inducing, and the sparsity vanishes when regularization is removed. We experimentally observe the following:
\vspace{-0.2cm}
\begin{packed_itemize}
\item The sparsity is much higher when using adaptive flavors of SGD vs. (m)SGD.
\item Adaptive methods see higher sparsity with L2 regularization than with WD. No sparsity emerges in the absence of regularization.
\item In addition to the regularizers, the extent of the emergent sparsity is also influenced by hyperparameters seemingly unrelated to regularization. The sparsity decreases with increasing mini-batch size, decreasing network size and increasing task difficulty.
\item The primary hypothesis that we put forward is that selective features\footnote{Feature selectivity is the fraction of training exemplars for which a feature produces max activation less than some threshold.} see a disproportionately higher amount of regularization than non-selective ones. This consistently explains how parameters such as mini-batch size, network size, and task difficulty indirectly impact sparsity by affecting feature selectivity.
\item A secondary hypothesis to explain the higher sparsity observed with adaptive methods is that Adam (and possibly other) adaptive approaches learn more selective features. Though threre is evidence of highly selective features with Adam, this requires further study. 
\item  Synthetic experiments show that the interaction of L2 regularizer with the update equation in adaptive methods causes stronger regularization than WD. This can explain the discrepancy in sparsity between L2 and WD.
\vspace{-0.2cm}
\end{packed_itemize}

\parahead{Quantifying Feature Sparsity}
Feature sparsity can be measured by per-feature activation and by per-feature scale. For sparsity by activation, the absolute activations for each feature are max pooled over the entire feature plane. If the value is less than $10^{-12}$ over the entire \emph{training} corpus, the feature is inactive. For sparsity by scale, we consider the scale $\gamma$ of the learned affine transform in the Batch Norm layer. 
We consider a feature inactive if $|\gamma|$ for the feature is less than $10^{-3}$. Explicitly zeroing the features thus marked inactive does not affect the test error. 
The thresholds chosen are purposefully conservative, and comparable levels of sparsity are observed for a higher feature activation threshold of $10^{-4}$, and a higher $|\gamma|$ threshold of $10^{-2}$.

The implicit sparsification process can remove 70-80\% of the convolutional filters from VGG-16 on CIFAR10/100, far exceeding that for \cite{li2017pruning}, and performs comparable to \cite{liu2017learning} for VGG-11 on ImageNet. 
Common hyperparameters such as mini-batch size can be used as knobs to control the extent of sparsity, with no tooling changes needed to the traditional NN training pipeline.

We present some of the experimental results, and discuss links of the hypothesis to pruning heuristics.

\begin{figure}
\centering
  \includegraphics[width=0.80\columnwidth]{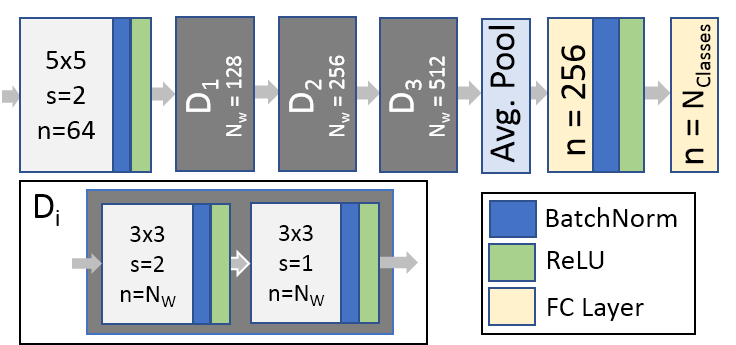}
  \vspace{-0.2cm}
  \caption{\textbf{BasicNet}: Structure of the basic convolution network studied in this paper. We refer to the convolution layers as C1-7.} 
  \label{fig:basic_net}
  \vspace{-0.5cm}
\end{figure}
\renewcommand{\tabcolsep}{1.5pt}
\begin{table}[ht!]
\centering
\caption{ Convolutional filter sparsity in \emph{BasicNet} trained on CIFAR10/100 for different combinations of regularization and gradient descent methods. Shown are the \% of non-useful / inactive convolution filters, as measured by activation over training corpus (max act. $< 10^{-12}$) and by the learned BatchNorm scale ($|\gamma| < 10^{-03}$), averaged over 3 runs. The lowest test error per optimizer is highlighted, and sparsity (green) or lack of sparsity (red) for the best and near best configurations indicated via text color. L2: L2 regularization, WD: Weight decay (adjusted with the same scaling schedule as the learning rate schedule).}
\label{tbl:all_optim}
\resizebox{0.75\linewidth}{!}{
\begin{tabular}{c|c|c|c|c|c|c|c|c|}
\multicolumn{9}{c}{}\\
\cline{2-9}
                              &        & \multicolumn{3}{c}{CIFAR10}               &  &   \multicolumn{3}{c|}{CIFAR100}                      \\ \cline{3-9}
                              &        & \multicolumn{2}{c|}{\% Sparsity}    & Test           &  &   \multicolumn{2}{c|}{\% Sparsity} & Test                     \\
                              & \textbf{L2}       & by Act & by $\gamma$   & Error         &   & by Act & by $\gamma$   & Error              \\ \hline
                               & 1e-03 & 27                           & 27                        & 21.8                                                                                  &  & 23                           & 23                        & 47.1                                                                               \\
                               & 1e-04 &  0     & 0  & 11.8                                                          &  &                {\color[HTML]{9A0000}0}                &           {\color[HTML]{9A0000}0}                  &           \cellcolor[HTML]{9AFF99}37.4                                                                                \\
                               & 1e-05 &  {\color[HTML]{9A0000} 0}     & {\color[HTML]{9A0000} 0}  & \cellcolor[HTML]{9AFF99}10.5                                                          &  &  0     & 0  & 39.0                                                         \\ 
\parbox[t]{2mm}{\multirow{-4}{*}{{\rotatebox[origin=c]{90}{SGD\B}}}}                              & 0        & 0                            & 0                         & 11.3                                                                                  &  & 0                            & 0                         & 40.1                                                                                  \\ \hline
                               & 2e-03 & 88                           & 86                        & 14.7                                                                                 &  & 82                           & 81                        & 42.7                                                                                 \\
                               & 1e-04 & {\color[HTML]{009901} 71}    & {\color[HTML]{009901} 70} & \cellcolor[HTML]{9AFF99}10.5                                                          &  & {\color[HTML]{009901} 47}    & {\color[HTML]{009901} 47} & \cellcolor[HTML]{9AFF99}36.6                                                          \\
                               & 1e-05 & {\color[HTML]{009901} 48}    & {\color[HTML]{009901} 48} & 10.7                                                          &  & {\color[HTML]{9A0000} 5}                            & {\color[HTML]{9A0000} 5}                         & 40.6                                                                                  \\
 \multirow{-4}{*}{{\rotatebox[origin=c]{90}{Adam}}}                              & 0        & 3                            & 0                         & 11.0                                                                                  &  & 0                            & 0                         & 40.3                                                                                  \\\hline
                               & \Te 5e-04 &  82     &  82  & 13.6                                                          &  & {\color[HTML]{009901}61}                            & {\color[HTML]{009901}61}                         & 39.1                                                          \\
                               & 2e-04 & {\color[HTML]{009901} 40}     & {\color[HTML]{009901} 40}  & 11.3                                                          &  & {\color[HTML]{9A0000}3}                            & {\color[HTML]{9A0000}3}                         & \cellcolor[HTML]{9AFF99}35.4                                                          \\
\multirow{-3}{*}{{\rotatebox[origin=c]{90}{Adadelta}}}         & \Be 1e-04 & {\color[HTML]{9A0000} 1}     & {\color[HTML]{9A0000} 1}  & \cellcolor[HTML]{9AFF99}10.2                                                          &  & 1                            & 1                         & 35.9                                                          \\ \hline
                               & \Te 2e-02 & {\color[HTML]{009901} 75}    & {\color[HTML]{009901} 75} & 11.3                                                          &  & 88                           & 88                        & 63.3                                                          \\
                               & 1e-02 & {\color[HTML]{009901} 65}    & {\color[HTML]{009901} 65} & \cellcolor[HTML]{9AFF99}11.2                                                          &  & {\color[HTML]{009901}59}                           & {\color[HTML]{009901}59}                        & 37.2                                                          \\
\multirow{-3}{*}{{\rotatebox[origin=c]{90}{Adagrad}}}                               & \Be 5e-03 & {\color[HTML]{009901} 56}    & {\color[HTML]{009901} 56} & 11.3                                                          &  & {\color[HTML]{009901}24}                           & {\color[HTML]{009901}25}                        & \cellcolor[HTML]{9AFF99}35.9                                                          \\ \hline
                               & \Te 1e-02  & 93                          & 93                       & 20.9                                                                                  &  & 95                          & 95                       & 71.9                                                                                  \\
                               & 1e-04 & 51                           & 47                        & 9.9                                                                                  &  & 20                           & 13                        & 35.6                                                                                  \\
\multirow{-3}{*}{{\rotatebox[origin=c]{90}{AMSGrad}}}         & \Be 1e-06 &  0     &  0  & 11.2                                                          &  & {0 }                           & {0}                         & 40.2                                                                                  \\ \hline 
                               & \Te 1e-02  & 75                          & 90                       & 16.4                                                                                  &  & 74                          & 87                      & 51.8                                                                                  \\
                               & 1e-04 & 49                           & 50                        & 10.1                                                                                  &  & 10                           & 10                        & 39.3                                                                                  \\
\multirow{-3}{*}{{\rotatebox[origin=c]{90}{Adamax}}}         & \Be 1e-06 &  4     &  4  & 11.3                                                          &  & {0 }                           & {0}                         & 39.8                                                                                  \\ \hline 
                               & \Te 1e-02  & 95                          & 95                       & 26.9                                                                                  &  & 97                          & 97                       & 78.6                                                                                  \\
                               & 1e-04 & 72                           & 72                        & 10.4                                                                                  &  & 48                           & 48                        & 36.3                                                                                  \\
\multirow{-3}{*}{{\rotatebox[origin=c]{90}{RMSProp}}}         & \Be 1e-06 &  29     &  29  & 10.9                                                          &  & {0 }                           & {0}                         & 40.6                                                                                  \\ \hline 
                              &        & \multicolumn{3}{c}{CIFAR10}               &  &   \multicolumn{3}{c|}{CIFAR100}                      \\ \cline{3-9}
                              &        & \multicolumn{2}{c|}{\% Sparsity}    & Test           &  &   \multicolumn{2}{c|}{\% Sparsity} & Test                     \\
                              & \textbf{WD}       & by Act & by $\gamma$   & Error         &   & by Act & by $\gamma$   & Error              \\ \hline
                               & 1e-03 & 27                           & 27                        & 21.6                                                                                  &  & 23                           & 23                        & 47.6                                                                                  \\
                               & 2e-04 & 0    & 0  & 13.3            &  & 0     & 0  & 39.4                                  \\
\multirow{-3}{*}{{\rotatebox[origin=c]{90}{SGD}}}         & 1e-04 & {\color[HTML]{9A0000} 0}     & {\color[HTML]{9A0000} 0}  & \cellcolor[HTML]{9AFF99}12.4                                                          &  & {\color[HTML]{9A0000}0 }                           & {\color[HTML]{9A0000}0}                         & \cellcolor[HTML]{9AFF99}37.7                                                                                  \\ \hline 
                               & 5e-04 & 81                           & 81                        & 18.1                                                                                  &  &               59                 &      59                       &    43.3                                                                                       \\
                               & 2e-04 & {\color[HTML]{009901} 60}     & {\color[HTML]{009901} 60}  & 13.4                                                          &  & {\color[HTML]{009901} 16}     & {\color[HTML]{009901} 16}  &  37.3                                                          \\
\multirow{-3}{*}{{\rotatebox[origin=c]{90}{Adam}}}        & 1e-04 & {\color[HTML]{009901} 40}     & {\color[HTML]{009901} 40}  & \cellcolor[HTML]{9AFF99}11.2                                                          &  &    {\color[HTML]{9A0000}3}                            &        {\color[HTML]{9A0000}3}                     &   \cellcolor[HTML]{9AFF99}36.2                                                                                  \\ \hline
\end{tabular}
}
\vspace{-0.2cm}
\end{table}

\section{Related Work}
Among the various explicit filter level sparsification heuristics and approaches ~\cite{li2017pruning,srinivas2015data,hu2016network,theis2018faster,molchanov2017pruning,mozer1989skeletonization,liu2017learning,ye2018rethinking}, some \cite{ye2018rethinking,liu2017learning} make use of the learned scale parameter $\gamma$ in Batch Norm for enforcing sparsity on the filters. 
\cite{ye2018rethinking} argue that BatchNorm makes feature importance less susceptible to scaling reparameterization, and the learned scale parameters ($\gamma$) can be used as indicators of feature importance.
We find that Adam with L2 regularization, owing to its implicit pruning of features based on feature selectivity, makes it an attractive alternative to explicit sparsification/pruning approaches. The link between ablation of selective features and explicit feature pruning is also established in prior work~\cite{morcos2018importance,zhou2018revisiting}. 
We show that the emergence of selective features in Adam, and the increased propensity for pruning the said selective features when using L2 regularization may thus be helpful both for better generalization performance and network speedup.

\section{Experimental Setup and Results}
\label{sec:basic_setup}
We use a 7-layer convolutional network we with 2 fully connected layers as shown in Figure~\ref{fig:basic_net}. We refer to this network as \emph{BasicNet} in the rest of the document. We also compare VGG-11/16 \cite{simonyan2014very} on various datasets. 
For the basic experiments on CIFAR-10/100, we use a variety of gradient descent approaches, a mini-batch size of 40, with a method specific base learning rate for 250 epochs which is scaled down by 10 for 75 more epochs. The base learning rates and other hyperparameters are as follows: Adam (1e-3, $\beta_1$=0.9, $\beta_2$=0.99, $\epsilon$=1e-8), Adadelta (1.0, $\rho$=0.9, $\epsilon$=1e-6), SGD (0.1, mom.=0.9), Adagrad (1e-2), AMSGrad (1e-3), AdaMax (2e-3), RMSProp (1e-3).  

\parahead{Observing Sparsity Trends}
We study the effect of varying the amount and type of regularization\footnote{Note that L2 regularization and weight decay are distinct. See \cite{loshchilov2017fixing} for a detailed discussion.} on the extent of sparsity and test error in Table \ref{tbl:all_optim}. It shows significant convolutional filter sparsity emerges with adaptive gradient descent methods when combined with L2 regularization. The extent of sparsity is reduced when using Weight Decay instead, and absent entirely in the case of SGD with moderate levels of regularization.
The extent of sparsity decreases with increasing mini-batch size, as shown in Tables~\ref{tbl:bat_size},~\ref{tbl:tinyimagenet_basic},~\ref{tbl:vgg_cifar},~\ref{tbl:vgg16_tinyimagenet},~\ref{tbl:vgg11_imagenet} and~\ref{tbl:vgg19_cifar}.
%
\begin{table}[htb!]
\vspace{-0.2cm}
\renewcommand{\tabcolsep}{1.5pt}
\centering
\caption{BasicNet sparsity variation on CIFAR10/100 trained with Adam and L2 regularization with different mini-batch sizes.}
\label{tbl:bat_size}
\resizebox{0.85\linewidth}{!}{
\begin{tabular}{cc|c|c|c|c|c|cccc}
\multicolumn{1}{l}{}                             & \multicolumn{1}{l|}{} & \multicolumn{4}{c|}{CIFAR 10}                                                  & \multicolumn{1}{l|}{} & \multicolumn{4}{c}{CIFAR 100}                                                                                                                 \\
\multicolumn{1}{l|}{}                            & Batch                 & Train                       & Test                        & Test & \%Spar.  &                       & \multicolumn{1}{c|}{Train}                       & \multicolumn{1}{c|}{Test}                        & \multicolumn{1}{c|}{Test} & \%Spar.  \\
\multicolumn{1}{c|}{}                            & Size                  & Loss                        & Loss                        & Err  & by $\gamma$ &                       & \multicolumn{1}{c|}{Loss}                        & \multicolumn{1}{c|}{Loss}                        & \multicolumn{1}{c|}{Err}  & by $\gamma$ \\ \hline
\multicolumn{1}{c|}{}                           & 20                    & {\color[HTML]{9B9B9B} 0.17} & {\color[HTML]{9B9B9B} 0.36} & 11.1 & 70          &                       & \multicolumn{1}{c|}{{\color[HTML]{9B9B9B} 0.69}} & \multicolumn{1}{c|}{{\color[HTML]{9B9B9B} 1.39}} & \multicolumn{1}{c|}{35.2} & 57          \\
\multicolumn{1}{c|}{}                           & 40                    & {\color[HTML]{9B9B9B} 0.06} & {\color[HTML]{9B9B9B} 0.43} & 10.5 & 70          &                       & \multicolumn{1}{c|}{{\color[HTML]{9B9B9B} 0.10}} & \multicolumn{1}{c|}{{\color[HTML]{9B9B9B} 1.98}} & \multicolumn{1}{c|}{36.6} & 46          \\
\multicolumn{1}{c|}{}                           & 80                    & {\color[HTML]{9B9B9B} 0.02} & {\color[HTML]{9B9B9B} 0.50} & 10.1 & 66          &                       & \multicolumn{1}{c|}{{\color[HTML]{9B9B9B} 0.02}} & \multicolumn{1}{c|}{{\color[HTML]{9B9B9B} 2.21}} & \multicolumn{1}{c|}{41.1} & 35          \\
\multicolumn{1}{c|}{\multirow{-4}{*}{\rotatebox[origin=c]{90}{L2: 1e-4\B}}} & 160                   & {\color[HTML]{9B9B9B} 0.01} & {\color[HTML]{9B9B9B} 0.55} & 10.6 & 61          &                       & \multicolumn{1}{c|}{{\color[HTML]{9B9B9B} 0.01}} & \multicolumn{1}{c|}{{\color[HTML]{9B9B9B} 2.32}} & \multicolumn{1}{c|}{44.3} & 29          \\ \hline
\end{tabular}
}
\label{tbl:bat_size}
\vspace{-0.4cm}
\end{table}
\begin{table}[ht!]
\vspace{-0.2cm}
\renewcommand{\tabcolsep}{1.5pt}
\centering
\caption{Convolutional filter sparsity for BasicNet trained on TinyImageNet, with different mini-batch sizes. 
}
\label{tbl:tinyimagenet_basic}
\resizebox{0.7\linewidth}{!}{
\begin{tabular}{c|c|c|c|c|c|c}
                       & Batch & Train                       & Val                         & Top 1    & Top 5    & \% Spar. \\
                       & Size  & Loss                        & Loss                        & Val Err. & Val Err. & by $\gamma$ \\ \hline
                       & 20    & {\color[HTML]{9B9B9B} 1.05} & {\color[HTML]{9B9B9B} 2.13} & 47.7     & 22.8     & 63          \\
                       & 40    & {\color[HTML]{9B9B9B} 0.16} & {\color[HTML]{9B9B9B} 2.96} & 48.4     & 24.7     & 48          \\
 & 120   & {\color[HTML]{9B9B9B} 0.01} & {\color[HTML]{9B9B9B} 2.48} & 48.8     & 27.4     & 26          \\ \hline
\end{tabular}
}
\label{tbl:tnyimngnet_basic}
\vspace{-0.4cm}
\end{table}
\begin{table}[htb!]
\renewcommand{\tabcolsep}{1.5pt}
\centering
\caption{Layerwise \% Sparsity by $\gamma$ for VGG-16 on CIFAR10 and 100, with different mini-batch sizes, compared to the handcrafted sparse structure of \cite{li2017pruning}}
\resizebox{0.9\linewidth}{!}{
\begin{tabular}{cc|c|c|c|c|c|ccc}
                              &            & \multicolumn{4}{c|}{CIFAR 10}                                              &  & \multicolumn{3}{c}{CIFAR 100}               \\ \cline{3-10}
\multicolumn{1}{c}{}     &            & \multicolumn{3}{c|}{Adam, L2:1e-4}          &      {\color[HTML]{000199}}                      &  & \multicolumn{3}{c}{Adam, L2:1e-4}          \\
\multicolumn{1}{c}{}    &  & B: 40 & B: 80 & B: 160                     & {\color[HTML]{000199}Li et al.}         &  & \multicolumn{1}{c|}{B: 40}     & \multicolumn{1}{c|}{B: 80} & B: 160    \\ \hline
\multicolumn{2}{c|}{\%Feat. Pruned}                & 86     & 84    & {\color[HTML]{009901}83}  & {\color[HTML]{009901}37}  &  & \multicolumn{1}{c|}{76}         & 69   &  \multicolumn{1}{|c}{69}     \\ \hline \hline
\multicolumn{2}{c|}{Test Err}             & 7.2    & 7.0   & \cellcolor[HTML]{9AFF99}6.5 & \cellcolor[HTML]{9AFF99}6.6 &  & \multicolumn{1}{c|}{29.2}       & 28.1 &   \multicolumn{1}{|c}{27.8}   
\end{tabular}
}

\label{tbl:vgg_cifar}
\vspace{-0.4cm}
\end{table}
\begin{table}[ht!]
\renewcommand{\tabcolsep}{2.5pt}
\centering
\caption{Sparsity by $\gamma$ on VGG-16, trained on TinyImageNet, and on ImageNet. Also shown are the pre- and post-pruning top-1/top-5 single crop validation errors. Pruning using $|\gamma| < 10^{-3}$ criteria.}
\label{tbl:vgg16_tinyimagenet}
\resizebox{0.95\linewidth}{!}{
\begin{tabular}{c|c|cc|c|c|}
             &        \# Conv               & \multicolumn{2}{c|}{Pre-pruning}    & \multicolumn{2}{c|}{Post-pruning} \\
TinyImageNet                &  Feat. Pruned & \multicolumn{1}{c|}{top1}  & top5  & top1          & top5        \\ \hline
L2: 1e-4, B: 20 & 3016 (71\%)           & \multicolumn{1}{c|}{45.1} & 21.4 & 45.1         &     21.4       \\
L2: 1e-4, B: 40 & 2571 (61\%)           & \multicolumn{1}{c|}{46.7} & 24.4 & 46.7         &     24.4       \\ \hline
 ImageNet    &       \multicolumn{5}{c|}{} \\ \hline
L2: 1e-4, B: 40 & 292           & \multicolumn{1}{c|}{29.93} & 10.41 & 29.91         &     10.41       \\
\end{tabular}
}
\label{tbl:vgg16_tinyimagenet}
\vspace{-0.2cm}
\end{table}

\parahead{Comparison With Explicit Feature Sparsification Approaches}
For VGG-16, we compare the network trained on CIFAR-10 with Adam using different mini-batch sizes against the handcrafted approach of Li et al.~\cite{li2017pruning}. Similar to tuning the explicit sparsification hyperparameter in~\cite{liu2017learning}, the mini-batch size can be varied to find the sparsest representation with an acceptable level of test performance. We see from Table~\ref{tbl:vgg_cifar} that when trained with a batch size of 160, 83\% of the features can be pruned away and leads to a better performance that the 37\% of the features pruned for ~\cite{li2017pruning}.
For VGG-11 on ImageNet (Table~\ref{tbl:vgg11_imagenet}), by simply varying the mini-batch size from 90 to 60, the number of convolutional features pruned goes from 71 to 140. This is in the same range as the number of features pruned by the explicit sparsification approach of ~\cite{li2017pruning}, and gives a comparable top-1 and top-5 validation error. 
For VGG-19 on CIFAR10 and CIFAR100 (Table~\ref{tbl:vgg19_cifar}), we see again that varying the mini-batch size controls the extent of sparsity. For the mini-batch sizes we considered, the extent of sparsity is much higher than that of \cite{liu2017learning}, with consequently slightly worse performance. Tweaking the mini-batch size or other hyper-parameters would further tradeoff sparsity for accuracy, and reach a comparable sparsity-accuracy as~\cite{liu2017learning}. 

\begin{table}[htb!]
\vspace{-0.2cm}
\renewcommand{\tabcolsep}{2.5pt}
\centering
\caption{Effect of different mini-batch sizes on sparsity (by $\gamma$) in VGG-11, trained on ImageNet. Same network structure employed as \cite{liu2017learning}. * indicates finetuning after pruning}
\label{tbl:vgg11_imagenet}
\resizebox{\linewidth}{!}{
\begin{tabular}{l|c|cc|c|c|}
                &       \# Conv               & \multicolumn{2}{c|}{Pre-pruning}    & \multicolumn{2}{c|}{Post-pruning} \\
                &  Feat. Pruned & \multicolumn{1}{c|}{top1}  & top5  & top1          & top5        \\ \hline
Adam, L2: 1e-4, B: 90 & {\color[HTML]{009901}71}                     & \multicolumn{1}{c|}{30.50} & 10.65 & \cellcolor[HTML]{9AFF99}30.47         &     10.64       \\
Adam, L2: 1e-4, B: 60 & {\color[HTML]{009901}140}                    & \multicolumn{1}{c|}{31.76} & 11.53 & 31.73         &     11.51        \\ \hline
{\color[HTML]{000199}\cite{liu2017learning}}      & {\color[HTML]{009901}85}                     & \multicolumn{1}{c|}{29.16} &       & 31.38*         &   -         
\end{tabular}
}
\label{tbl:vgg11_imagenet}
\vspace{-0.4cm}
\end{table}
\begin{table}[htb!]
\renewcommand{\tabcolsep}{1.5pt}
\centering
\caption{Sparsity by $\gamma$ on VGG-19, trained on CIFAR10/100. Also shown are the post-pruning test error. Compared with explicit sparsification approach of \cite{liu2017learning} }
\label{tbl:vgg19_cifar}
\resizebox{\linewidth}{!}{
\begin{tabular}{c|c|c|c|c|ccc}
                                          & \multicolumn{3}{c|}{CIFAR 10}                                              &  & \multicolumn{3}{c}{CIFAR 100}               \\ \cline{1-8}
\multicolumn{1}{c|}{}            & \multicolumn{2}{c|}{Adam, L2:1e-4}          &      {\color[HTML]{000199}}                       &  & \multicolumn{2}{c}{Adam, L2:1e-4} & {\color[HTML]{000199} }       \\
\multicolumn{1}{c|}{}     & B: 64 & B: 512&      {\color[HTML]{000199}\cite{liu2017learning}}          &  & \multicolumn{1}{c|}{B: 64}     & \multicolumn{1}{c|}{B: 512} &  {\color[HTML]{000199}\cite{liu2017learning}}    \\ \hline
\multicolumn{1}{c|}{\%Feat. Pruned}                & 85     & {\color[HTML]{009901}81}    & {\color[HTML]{009901}70}    &  & \multicolumn{1}{c|}{75}         & {\color[HTML]{009901}62}   &  \multicolumn{1}{|c}{\color[HTML]{009901}50}     \\ \hline \hline
\multicolumn{1}{c|}{Test Err}             & 7.1    & 6.9   & \cellcolor[HTML]{9AFF99}6.3  &  & \multicolumn{1}{c|}{29.9}       & 28.8 &   \multicolumn{1}{|c}{\cellcolor[HTML]{9AFF99}26.7}   
\end{tabular}
}
\label{tbl:vgg19_cifar}
\vspace{-0.5cm}
\end{table}
\section{Discussion}
The penalization of selective features can be seen as a greedy local search heuristic. 
While the extent of implicit filter sparsity is significant, it obviously does not match up with some of the more recent approaches~\cite{he2018amc,lin_rnp_nips2017} which utilize more expensive model search and advanced heuristics such as filter redundancy.

One question that emerges from our findings is whether other filter sparsification heuristics can be implicitly incorporated into the network learning process, such that the training process jointly optimizes for the training objective and network capacity.

Another issue to consider is the selective-feature pruning criteria itself. On the opposite end of the spectrum from selective features lie non-selective features, which putatively have comparably low discriminative information as selective features and could also be pruned. These non-selective features are however not captured by greedy local search heuristics because pruning them can have a significant impact on the accuracy. The accuracy can presumably can be recouped after fine-tuning, and this remains open as a possible line of future investigation.


\bibliography{article}
\bibliographystyle{icml2019}
\end{document}